\documentclass[final,5p,times,twocolumn]{article}
\usepackage{natbib}
\usepackage{url}
\usepackage{graphicx}

\begin{document}

\title{Identifying Bias in Deep Neural Networks Using Image Transforms}   

\author{Sai Teja Erukude, Akhil Joshi and Lior Shamir \\ Kansas State University \\ 1701 Platt St, Manhattan, KS 66506, USA}

\date{}
\maketitle



\abstract{
CNNs have become one of the most commonly used computational tool in the past two decades. 
One of the primary downsides of CNNs is that they work as a ``black box", where the user cannot necessarily know how the image data are analyzed, and therefore needs to rely on empirical evaluation to test the efficacy of a trained CNN. This can lead to hidden biases that affect the performance evaluation of neural networks, but are difficult to identify. Here we discuss examples of such hidden biases in common and widely used benchmark datasets, and propose techniques for identifying dataset biases that can affect the standard performance evaluation metrics. One effective approach to identify dataset bias is to perform image classification by using merely blank background parts of the original images. However, in some situations a blank background in the images is not available, making it more difficult to separate foreground or contextual information from the bias. 
To overcome this, we propose a method to identify dataset bias without the need to crop background information from the images. That method is based on applying several image transforms to the original images, including Fourier transform, wavelet transforms, median filter, and their combinations. These transforms were applied to recover background bias information that CNNs use to classify images. This transformations affect the contextual visual information in a different manner than it affects the systemic background bias. Therefore, the method can distinguish between contextual information and the bias, and alert on the presence of background bias even without the need to separate sub-images parts from the blank background of the original images. Code used in the experiments is publicly available.
}





\section{Introduction}
\label{introduction}

In the past decade, CNNs have revolutionized the field of computer vision due to their unprecedented performance capabilities with regard to image analysis, combined with ease-of-use due to the availability of libraries. These networks use automatic extraction and selection of data-driven abstract features, which means a substantial reduction in the need for manual feature engineering. CNNs are currently widely used in a variety of applications, including object detection, recognition, medical imaging, and many more. 

With the substantial advantages, CNNs also have notable limitations. For example, training CNNs is computationally expensive, normally involving the need for advanced GPUs or computing clusters for execution. Training CNNs often requires a large number of samples \citep{uchida2016further}, and overfitting is a common situation when using CNNs \citep{santos2022avoiding,thanapol2020reducing,gavrilov2018preventing,sikha2024deep}. The need for large training sets leads to the common use of data augmentation, which can in itself affect the bias in a controlled or uncontrolled manner \citep{pastaltzidis2022data,babul2024synthetic,raja2024synthetic}. Changing the size of the input images can also change the performance of a CNN \citep{richter2021input}. 

In addition to these limitations, one of the primary downsides of CNNs is their ``black box" nature. While CNNs are trained with data samples, the process of updating their weights through the training remains opaque. The rules that define how classifications are made are convoluted and unintuitive, and it has been proven difficult to interpret the reason behind the decision made by a deep neural network \citep{ball2023}. The definition of what a CNN really ``learns" from the data thus becomes equally difficult to make. In other words, lack of interpretability simply means that CNNs should be used with caution as their classification processes are not necessarily clear to the user \citep{9154417, make3040048}.

Ideally, a CNN that can achieve high accuracy in classification on benchmark datasets should perform well also in real-world settings. On the other hand, empirical studies \citep{ball2023, TommasiPCT15, sanchari2021} show that many popular datasets were biased, and possibly not representative of a model's performance concerning real-world image recognition. That shows that CNNs trained on benchmark datasets might be driven by information that is not limited to the information related to the task they aim at solving.

Tools that are used to identify biases or irrelevant information that drives the classification accuracy include saliency maps \citep{alqaraawi2020evaluating,kim2019saliency,arun2021assessing}, which can be used to visualize the signal through the spacial domain. That allows to inspect and identify sources of signal that might not be relevant to the computer vision problem at hand \citep{simonyan2013deep,pfau2019global}. The domain of adversarial neural networks has also been studied in the context of irrelevant information that can skew the way CNNs work, yet with no direct link to the relevant information \citep{hashimoto2020multi,lin2020adversarial}. Some solutions has also been proposed to detect adversarial samples \citep{wang2021smsnet,pertigkiozoglou2018detecting}. The need of deep neural networks for large datasets led to the practice of data augmentation. Data augmentation has been used to correct for a variety of biases in datasets, especially biases related to the distribution of the data and fairness of the analysis \citep{jaipuria2020deflating,sharma2020data,mclaughlin2015data,iosifidis2018dealing}.

As the use of CNNs in various fields continues to grow, it is imperative to address these limitations, which also include the risk of classification bias. That reinforces the need to develop methods that can assist researchers who develop CNNs, and allow them to reduce the risk of unknown or unexpected biases.

\section{Classification bias in benchmark datasets}
\label{classification_bias}

The term ``bias" in the context of benchmark datasets for image classification has been used to describe numerous situations. Those may include unbalanced distribution of the samples in the classes, biases in the labeling of the images, biased selection of data, and more \citep{torralba2011unbiased}. Unbalanced distribution of the samples in datasets is a well-known bias driven by an unequal number of samples in different classes. Machine learning can be biased by the number of samples a machine learning model is trained by, making more populated classes become more frequent in the classification outcomes. That leads to some classes with higher confusion rate than other classes, consequently leading to problems in the fairness and discrimination by machine learning algorithms. One of the common solutions to the problem is by balancing the classes through data augmentation \citep{jaipuria2020deflating,sharma2020data,mclaughlin2015data,iosifidis2018dealing}.   

Labeling bias is a situation of incorrect or inconsistent labeling. It is related to contextual information bias, in which elements that are not the target objects are present in one class more frequently than in other classes, and can therefore be used by a classifier to identify the class. Such biases can be detected by manual or automatic inspection of the contextual information, and also by comparing the performance when training with one dataset and testing with another dataset \citep{torralba2011unbiased}. For instance, if the training and testing with two different datasets leads to weaker classification accuracy compared to training and testing with the same dataset, it can be considered an indication of bias \citep{torralba2011unbiased}.

While these biases can be identified, one of the more challenging biases is driven by information that is not part of the visible visual content of the images, yet its presence in the images allows machine learning algorithms to classify the images correctly. Because the source of such bias is not necessarily known, identifying it becomes more challenging, and can deceive even experienced researchers.

Because it is difficult to identify, such classification bias is present in many widely used benchmark datasets \citep{shamir2008evaluation,model2015,majeed2020issues,TommasiPCT15,sanchari2021,DHAR2022100545,ball2023}. For instance, \citet{model2015} conducted an in-depth study comparing dataset bias across various traditional object recognition benchmarks, including COIL-20, COIL-100, NEC Animals, Caltech 101, etc. They employed a method that involved isolating a small, seemingly blank background area from each image - an area that does not contain any information about the object of interest. The findings revealed that all datasets that were tested still achieved classification accuracy above chance levels using these small sub-images, even though the sub-images lacked any visually interpretable information]. This technique was used to detect dataset bias in single-object recognition datasets, and to compare the extent of bias across different datasets \citep{model2015}.

A similar observation was made with common face recognition datasets \citep{shamir2008evaluation}. Experiments showed that by using just a small part of the background that does not include any part of the face or hair, the algorithms could still identify the faces with accuracy far higher than mere chance \citep{shamir2008evaluation}.

The impact of bias is substantially stronger in datasets of objects imaged in controlled environments \citep{model2015,sanchari2021}. Such datasets include commonly used benchmarks such as COIL-20, COIL-100, and NEC Animals. Datasets prepared from natural images collected from online sources include ImageNet \citep{imagenet}, among others. The fact that these datasets achieved classification accuracy well above mere chance using only seemingly uninformative background areas that seem identical to each other when observed with the naked human eye indicates the presence of artifacts or noise that may influence the classification. This potentially allows machine vision algorithms to classify images without properly identifying the objects, and therefore without solving the machine vision problem they intend to solve.

\citet{majeed2020issues} encountered similar challenges when they investigated the applicability of CNNs for detecting COVID-19 in chest X-ray images. Their study involved using 12 established CNN architectures in transfer learning mode across three publicly available chest X-ray databases. Additionally, they introduced a custom shallow CNN model, trained from scratch. For their experiments, chest X-ray images were input into CNN models without any pre-processing, mirroring how other studies have used these images.

To better understand how these CNNs made their predictions, they conducted a qualitative analysis using a method called Class Activation Mappings (CAM) \citep{winastwan2024cam}. CAMs allow researchers to generate heatmaps of the most discriminating regions and visualize which parts of the input image were most influential in the network's decision-making process. By mapping the activation from the CNN back to the original image, one can highlight class-specific regions that contributed most significantly to the model's classification, providing insights into which part of the image was the most relevant for detecting COVID-19.

\begin{figure*}[ht]
    \centering
    \includegraphics[scale=0.2]{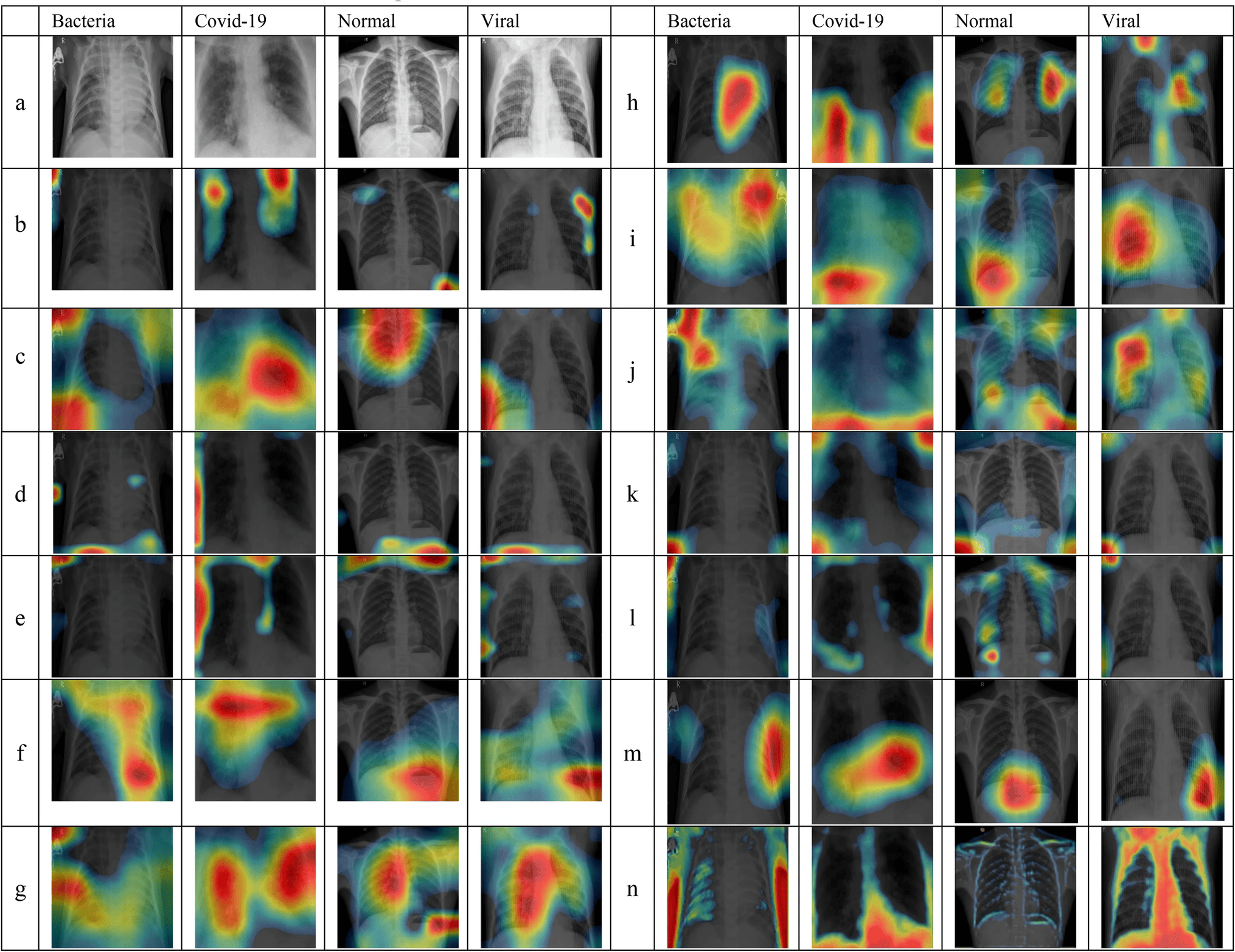}
    \caption{X-rays images classified correctly by CNNs. a) Original X-ray, b) AlexNet, c) GoogleNet, d) VGG16, e) VGG19, f) ResNet18, g) ResNet50, h) ResNet101, i) Inception V3, j) InceptionResNet, k) DenseNet201, l) SqueezeNet, m) Xception, and n) CNN-X \citep{majeed2020issues}}
    \label{fig_13models_cam}
\end{figure*}

Figure~\ref{fig_13models_cam} illustrates the parts of the images that the CNN models focused on when making predictions. As the figure shows, only in a few cases the models concentrated on the frontal chest region (i.e., the lung area), which is where typically signs of COVID-19 and other infections can be noticed. Instead, these models often focused on areas outside the chest's frontal view, and sometimes outside of the entire body. That can be seen in the first column of row (b) and the third and fourth columns of row (e) in Figure~\ref{fig_13models_cam}. Additionally, there are noticeable overlaps between the CAM hot spots and textual elements in the images, particularly in the first column of row (b), and the first, third, and fourth columns of row (e).

This is a notable example of how almost all CNN architectures used areas outside the region of interest (ROI) to make the final classification predictions. This observation leads to the conclusion that these images contain information that is not necessarily related to the medical condition being studied, yet can be used to classify between the images. The presence of such information can lead to biased classification accuracy such that the accuracy observed with the benchmark dataset is different from the classification accuracy that the CNN architecture can achieve in real-world settings.

The limitation of the method for the detection of biases is that the images are cannot be registered in all cases in a manner that allows a map that shows consistent signal from different parts of the images. Another limitation is that in some cases the entire image covers area from which signal is expected. In such cases, parts of the images that are known to be irrelevant cannot be inspected separately, and therefore the bias cannot be identified. When no parts that are irrelevant for the classification are expected, it is difficult to know whether bias exists, as all parts of the image are expected to be informative. 

\begin{figure}[ht]
    \centering
    \includegraphics[scale=1.0]{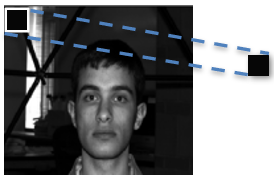}
    \caption{A 20$\times$20 cropped background segment of a blank background part of the original image. That was done to all images in the dataset to create a new dataset of blank sub-images, as shown in Figure~\ref{fig_yale_cropped}. When using just these seemingly blank parts of the images, the classification accuracy of numerous datasets was far higher than mere chance accuracy.}
    \label{fig_cropping_bg}
\end{figure}

A related study \citep{sanchari2021} highlighted significant bias-related issues in common biomedical, face recognition, and object recognition benchmark datasets. A primary concern is the tendency of CNNs to learn from background information rather than the relevant visual content of the images. The experiments follow common practices to test whether CNNs can classify these image datasets, but instead of using the entire images, merely parts of the images that do not have any content of the object of interest were used. An example is shown in Figure \ref{fig_cropping_bg}. 

\begin{figure*}[ht]
    \centering
    \includegraphics[scale=0.2]{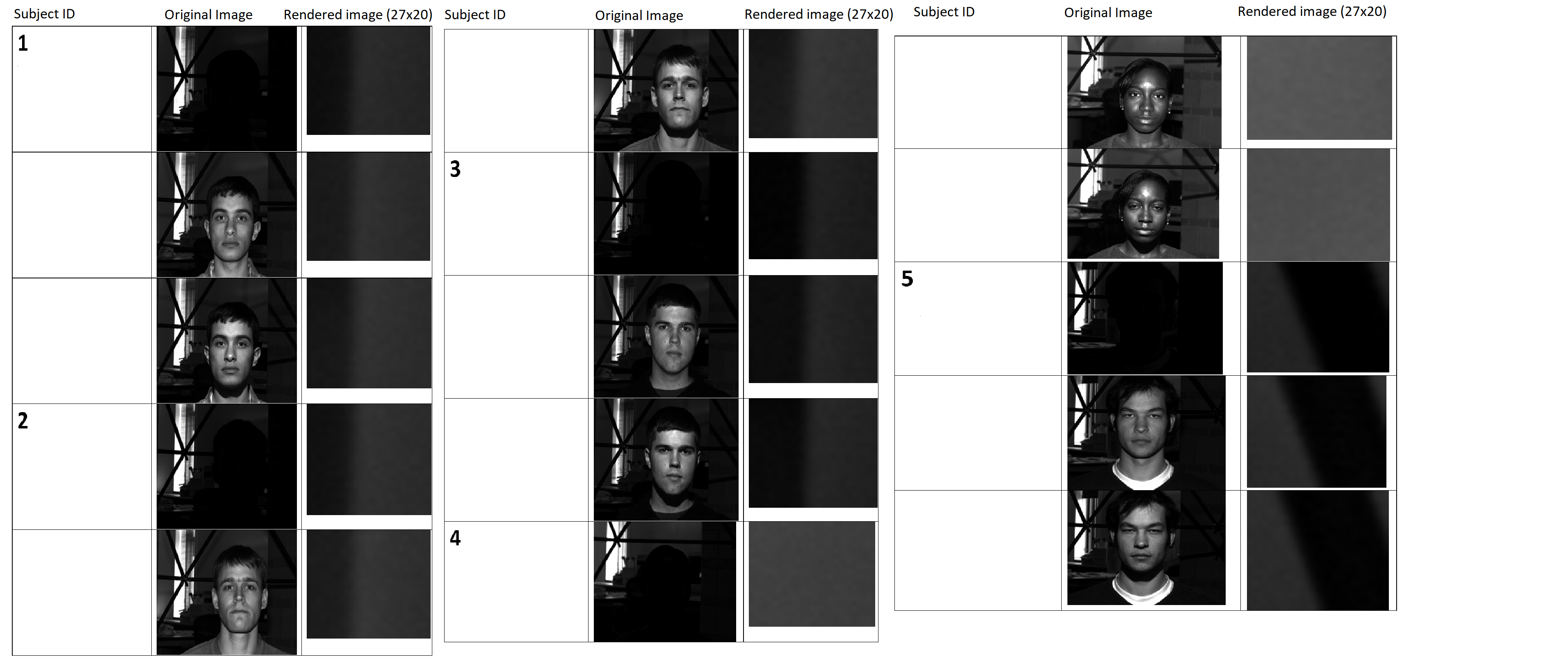}
    \caption{Original images from Yale Faces B and the 20$\times$20 portion of the top left corner separated from each of the original images. The classification accuracy of the CNN was far higher than mere chance, showing the the CNN does not necessarily need to recognize the face in order to classify the images correctly.}
    \label{fig_yale_cropped}
\end{figure*}

For instance, Figure \ref{fig_yale_cropped} depicts the original Yale Faces B dataset alongside the newly generated 20$\times$20 cropped dataset. The 20$\times$20 area of the image was selected based on previous experiments \citep{shamir2008evaluation,model2015,sanchari2021}, with the motivation of including just seemingly blank background parts of the images. Foreground parts of the images that contain information relevant to the image classification problem were not included in the sub-images. 

\begin{figure}[ht]
    \centering
    \includegraphics[scale=0.45]{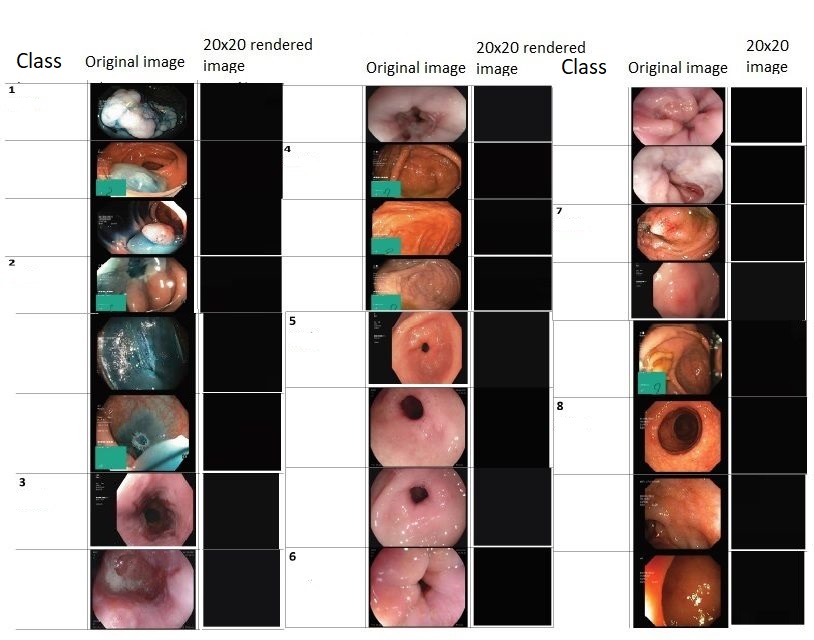}
    \caption{Original images from KVASIR and the 20$\times$20 portion of the top left corner separated from the original images \citep{sanchari2021}.}
    \label{fig_kvasir}
\end{figure}

When a CNN was trained on this cropped dataset, it achieved a surprisingly high classification accuracy of $\sim$87.8\%, which is far higher than the expected random accuracy of about $\sim$3.57\% \citep{sanchari2021}. Moreover, in the Kvasir dataset, which contains images of endoscopic examinations, CNNs were able to classify blank 20$\times$20 pixel sub-images as shown in Figure \ref{fig_kvasir}. In that case, the dataset made of small parts of the background provided classification accuracy of $\sim$30.75\%. That accuracy is also significantly higher than the expected random chance accuracy of $\sim$12.5\%. This raises questions about the reliability of models trained on such datasets, as they may exploit biases introduced by the image acquisition process rather than genuinely discerning clinical conditions.

\begin{figure}[ht]
    \centering
    \includegraphics[scale=0.6]{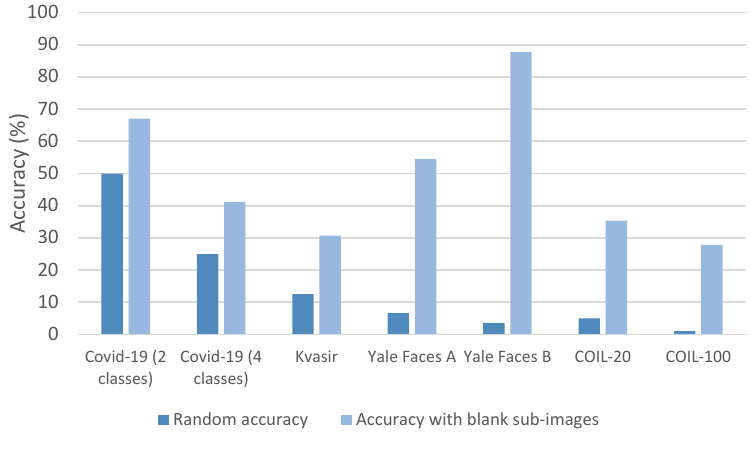}
    \caption{Classification accuracy of CNN models trained and tested on seemingly blank sub-images taken from image background of several common image benchmark datasets \citep{sanchari2021}. }
    \label{fig_sancharis_results}
\end{figure}

Experiments using various biomedical datasets, including COVID-CT and Kvasir, face recognition datasets such as Yale Faces A and Yale Faces B, and object recognition benchmarks such as COIL-20 and COIL-100, all showed that the classification accuracy of CNNs applied to these datasets might be biased \citep{sanchari2021}. In these studies, a small portion of the background was extracted to create new datasets, which were then utilized to train CNN models. The findings indicate that CNNs achieved high classification accuracy even when trained on datasets lacking relevant information or when systematically influenced by irrelevant features in the image data. The results are displayed in figure \ref{fig_sancharis_results}.

In summary, CNN prediction accuracy as determined by the standard practices should be treated cautiously, even when high accuracy is achieved, until the input image regions leading to those predictions are visually inspected. 
It is now evident that many dataset that are used commonly for machine learning experiments can contains some level of bias, and this should be taken into account before relying solely on high-accuracy figures. While the presence of such bias can be tested by using blank background from each image as done in \citep{model2015,sanchari2021}, in some cases background parts of the image might not be available. This paper proposes a method that involves the application of several image transforms, including Fourier, Wavelet transforms, and Median filter, and their combination. These transforms were applied to identify bias and recover background noise information used by CNN to classify images. The method is applied to the whole images, and can therefore be used in cases where no blank background is available to test for CNN bias.

\section{Data and CNN architectures}
\label{data}

\begin{figure}[ht]
    \centering
    \includegraphics[scale=0.4]{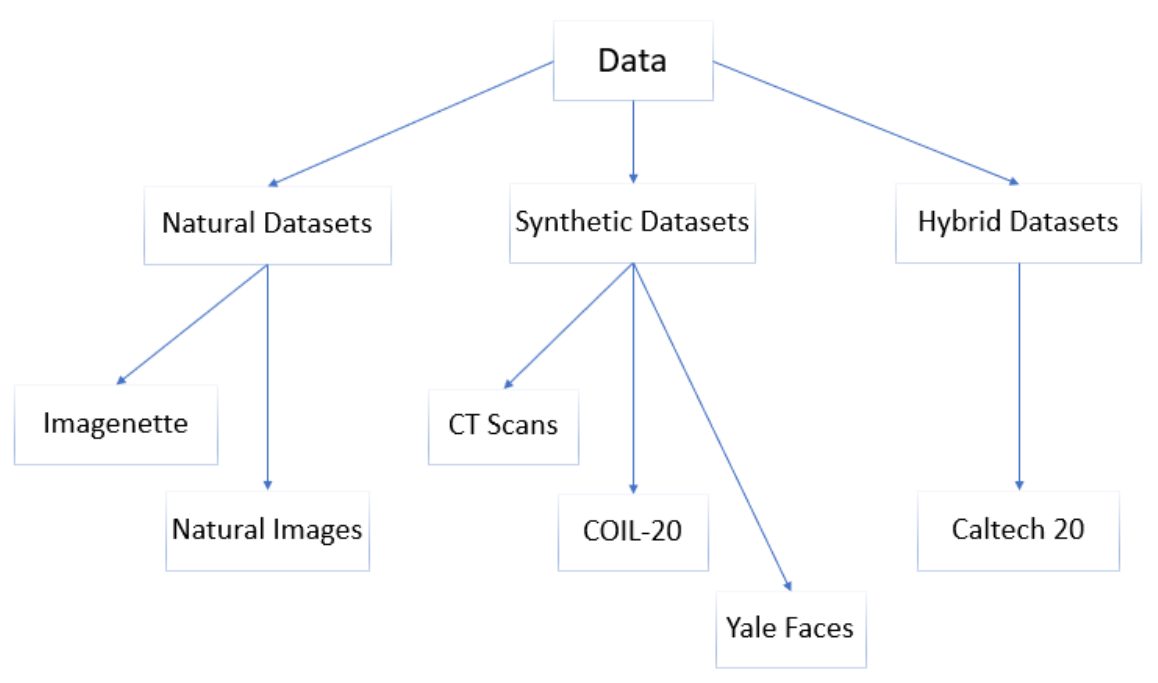}
    \caption{Categorization of data. The data used in this paper includes natural images collected from various sources, as well as controlled datasets such that all images in the dataset are provided from a single source.}
    \label{fig_data_categories}
\end{figure}

The experiments described in this paper were carried out across six distinct datasets to ensure that the findings were robust across different contexts. As shown in Figure~\ref{fig_data_categories}, datasets used in this study can be further categorized into three types: natural datasets, non-natural or synthetic datasets, and mixed or hybrid datasets.

\subsection{Natural datasets}
\label{natural_datasets}

Natural datasets comprise real-world images collected from a high number of different sources, and exhibit high variability in lighting, angles, and object appearances. Two such naturally collected datasets were employed in this study: \newline \newline 
\textit{Imagenette}: A commonly used smaller subset of the {\it ImageNet} \citep{imagenet} dataset, containing 10 different classes. The dataset contains images collected from a broad variety of sources, and is very commonly used in machine learning experiments. Imagenette is accessible at \url{https://github.com/fastai/imagenette}.
\newline \newline
\textit{Natural Images}: consists of images coming from eight different classes collected from several sources \citep{roy2018effects}. The classes include airplane, car, cat, dog, flower, fruit, motorbike and person. The class ``fruit" is discarded in this research because the background was changed to white, which could be used to identify that class and therefore can introduce biases in the classification process.

\subsection{Synthetic datasets}
\label{synthetic_datasets}

These datasets are either synthetically generated or created under controlled conditions, exhibiting less variability compared to natural datasets. Because of the lower variability and the controlled process, these datasets have higher vulnerability to bias when analyzed by machine learning \citep{shamir2008evaluation,model2015,sanchari2021}. Therefore, these datasets can be used as datasets that are known as datasets that contain information leading to bias, and any method for bias identification needs to be sensitive to the bias in these datasets. Following are the three different synthetic datasets used: \newline \newline
\textit{CT Scans}: Contains lung CT scan images across four classes, which are {\it normal}, {\it COVID-19}, {\it bacterial pneumonia}, and {\it viral pneumonia}. The dataset has been shown to contain certain information that can significantly influence CNN's learning regardless of its medical relevance \citep{majeed2020issues}. \newline \newline
{\it COIL-20}: The Columbia Object Image Library \citep{nene1996columbia} is a repository of gray-scale pictures that contains such twenty objects. All objects could be precisely put on a turntable. The turntable was turned at 360 degrees until a change of object was appropriately visible from a static camera. Images of the objects were captured in 5-degree angular orientations. \citet{model2015} demonstrated that {\it COIL-20}, being generated in a controlled environment, contains biases that significantly affect the model's classification accuracy. The exceptionally high classification accuracy, even when the object was absent from the test image, means that the model learned patterns from bias or background that are irrelevant to the detection of the objects. The presence of such bias makes this dataset useful for the purpose of this study, aiming to identify bias in image datasets. \newline \newline
{\it Yale Faces}: Yale Face Database \citep{georghiades2001few, olga:2018} contains 165 images of 15 subjects, with different facial expressions and lighting conditions. It includes the facial expressions of happy and sad, with and without glasses, sleepy, normal, wink, and surprised. It also includes lighting from different directions. \citet{sanchari2021} assessed face recognition benchmark datasets, including {\it Yale Faces}, to evaluate CNN performance. Similarly to COIL-20, the findings reveal a consistent bias within the dataset, with the CNN achieving notably higher classification accuracy when only a part of the image was used, highlighting the significant extent of this bias.

\subsection{Mixed or hybrid datasets}
\label{mixed_datasets}

Hybrid datasets are collections of images that combine both real-world (natural) images and synthetically generated or modified images. The purpose of these datasets is to leverage the advantages of both types of data to improve the performance and generalization of models.

The {\it Caltech 20} database is a subset of the larger {\it Caltech 256} dataset \citep{griffin2007caltech}. This study used only the first 20 object categories to simplify the dataset. Interestingly, upon closer inspection, we found that the first 20 classes feature a combination of real-world images and either graphically generated or have had their backgrounds altered to a plain white/black backdrop. Therefore, {\it Caltech 20} is categorized as a mixed dataset throughout this study.

\subsection{CNN architecture}
\label{cnn_architecture}

\begin{figure}[ht]
    \centering
    \includegraphics[scale=0.1]{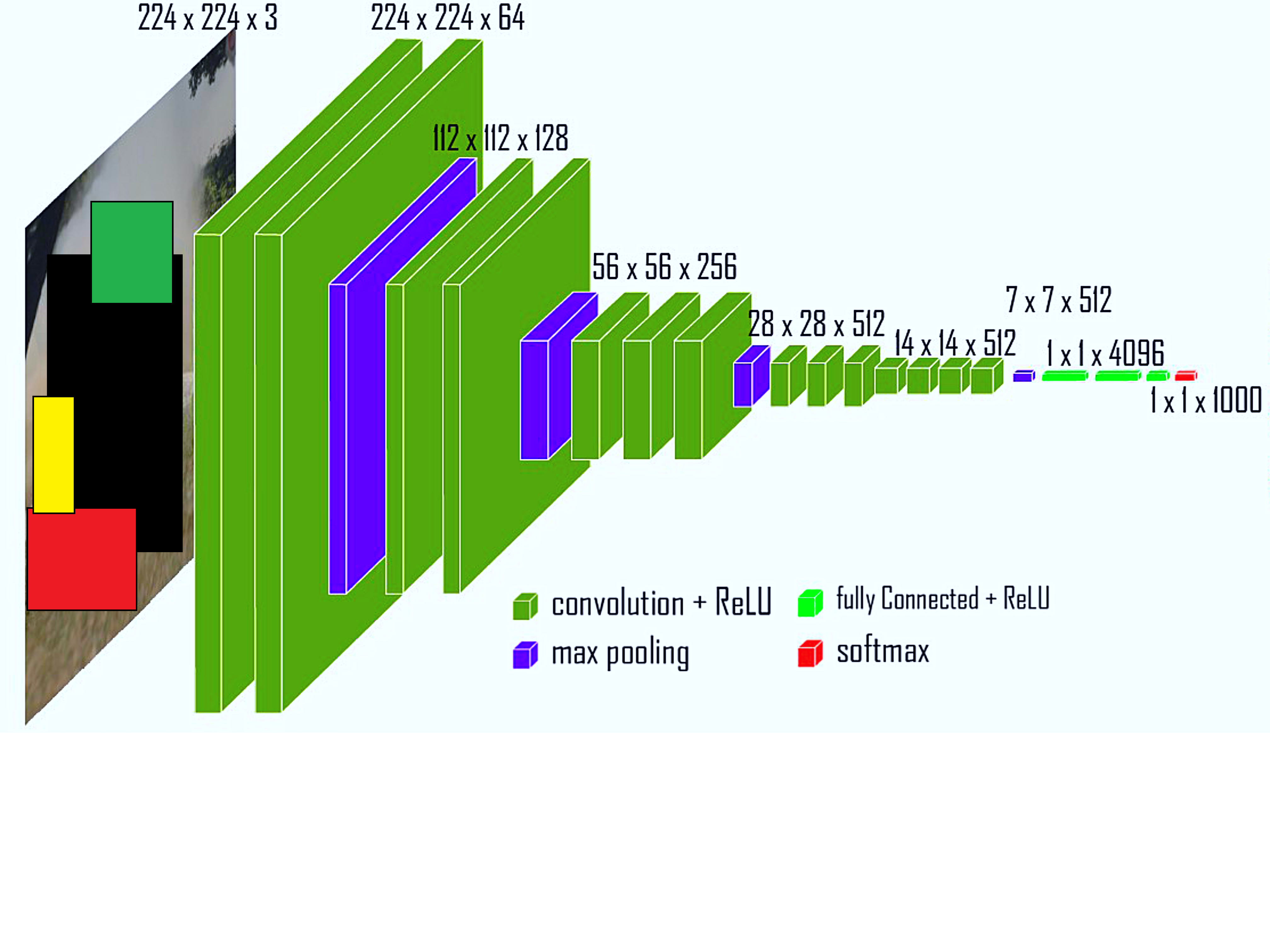}
    \caption{VGG16 architecture.}
    \label{fig_vgg16_architecture}
\end{figure}

If a certain CNN architecture can classify the images with background information alone, it is an indication that the dataset contains bias. Such bias introduce a risk that any CNN architecture tested on the data can be driven by that bias. This study utilized the very commonly used Visual Geometry Group 16 (VGG16) neural network architecture \citep{arXiv:1409.1556} that has demonstrated excellent classification accuracy. For instance, the architecture achieved high classification accuracy of $\sim$92.77\% on the full ImageNet dataset \citep{arXiv:1409.1556}. VGG16 comprises 13 convolutional layers, five max-pooling layers, and three fully connected (dense) layers, adding up to 21 layers. Only 16 of these layers have weights. The network takes as input images of size 224$\times$224 with three RGB channels. Figure~\ref{fig_vgg16_architecture} visualizes the VGG15 architecture. In all cases 70\% of the samples were allocated for the training set, 15\% for the test set, and 15\% for the validation set. The training was with 40 epochs and Adam optimizer. The code used for the experiments can be found at \url{https://github.com/SaiTeja-Erukude/identifying-bias-in-dnn-classification}. 

\section{Methods and results}
\label{methods_and_results}

As demonstrated in Section~\ref{classification_bias}, CNNs are capable of identifying the correct class even when they are trained and tested with seemingly blank background parts of the images. The presence of such bias can be identified by performing experiments similar to the experiments shown in Section~\ref{classification_bias}, repeating the same machine learning process using seemingly blank parts taken from the background of each image. But that approach can be used only if the images in the dataset has such blank background. If the entire image contains information that is meaningful for the classification, a control experiment with blank background cannot be performed, since better-than-random classification accuracy can be attributed to the visually relevant image content rather than the bias. Here we approach the task by applying image transforms to the original images, and testing the classification accuracy with the transformed images to identify the presence of biases.


\subsection{Fourier transform}
\label{fourier_transform}

Fourier transform is one of the most widely employed methods in image processing, used to break down an image into its sine and cosine frequency components \citep{cochran1967fast}. This transform converts the input spatial domain image to its equivalent frequency domain image. In the resulting image, each pixel denotes a certain frequency present in the spatial domain image. Since Discrete Fourier Transform (DFT) is a sampled version, it does not contain all frequencies forming an image. It is based only on a set of samples good enough to fully describe the image in the spatial domain. The number of frequencies equals the number of pixels in the spatial domain image, i.e. the image in the spatial and Fourier domains are of the same size \citep{fisher1996hypermedia}.





\begin{figure}[!ht]
    \centering
    \includegraphics[scale=0.6]{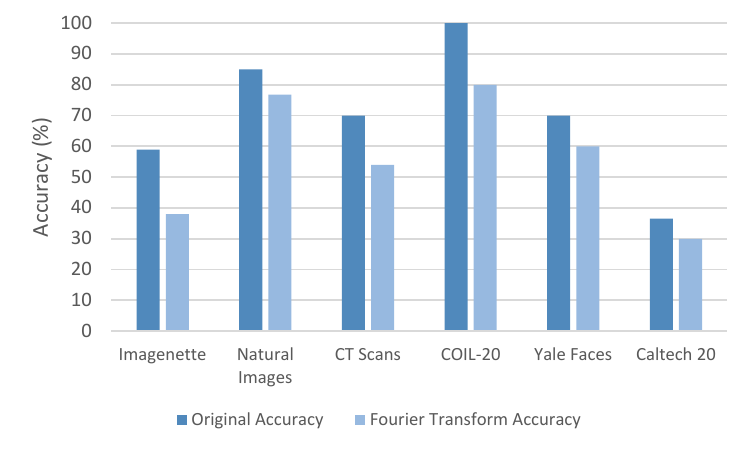}
    \caption{Classification accuracy when using the Fourier-transformed full images.}
    \label{fig_ft_full_graph}
\end{figure}

\begin{figure}[!ht]
    \centering
    \includegraphics[scale=0.6]{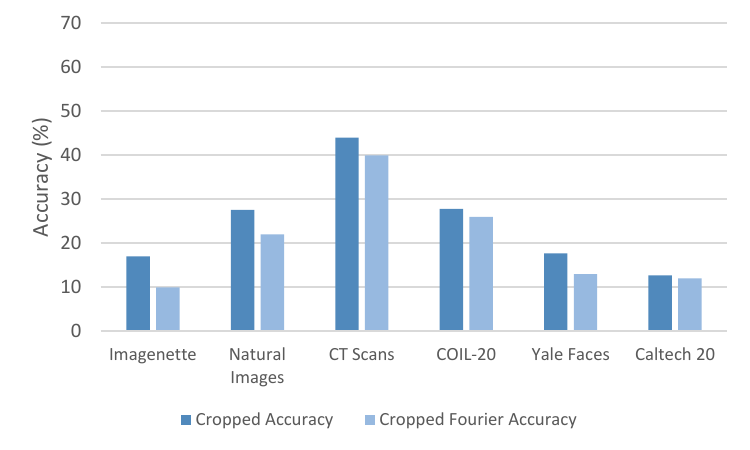}
    \caption{Classification accuracy when using the Fourier-transformed small sub-images taken from the background.}
    \label{fig_ft_cropped_graph}
\end{figure}

VGG16 models were trained and tested on the dataset after applying the Fourier transform to all images. Figures~\ref{fig_ft_full_graph} and~\ref{fig_ft_cropped_graph} show the classification accuracy of the full images after applying the Fourier transform, and the classification accuracy when using the 20$\times$20 sub-images cropped from the top right corner of each original image, and applying the Fourier transform. As the figures show, applying the Fourier transform has reduced the classification accuracy of the CNN across all the datasets, both for the full images and for the cropped sub-images. 

For example, in the Imagenette dataset, the accuracy dropped significantly from $\sim$59\% with the original full images to $\sim$38\% for Fourier-transformed full images. The accuracy for cropped sub-images dropped to around random levels. Similarly, the COIL-20 dataset saw its accuracy decline from $\sim$100\% to $\sim$80\%. As the figures show, most other datasets also showed a decrease in classification accuracy for both full and cropped images.

One possible explanation for this decrease could be that transforming images into the Fourier domain leads to a loss of spatial information as sensed by the VGG16 architecture. That is because VGG16 is not designed for the frequency domain, especially given that it is based on convolutional layers that are sensitive to pixels that are close to each other in the image. CNNs rely on this spatial information to effectively learn and classify features. Yet, using the Fourier-transformed images leaves it challenging to determine whether the drop in classification accuracy is due to the loss of relevant features or driven by the reduction of inherent bias in the datasets that were influencing the CNN's predictions. Therefore, the Fourier transform is an example of a transform that does not effectively help distinguish between biases in natural and non-natural datasets, as it reduces accuracy across all types of datasets.

\subsection{Wavelet transform}
\label{wavelet_transform}

Wavelet transforms are mathematical methods used for investigating and retrieving data from images \citep{agarwal2017analysis, othman2020applications}. They break an image into components at various positions, preserving both frequency and spatial data. In contrast to the Fourier transform, which outputs a global frequency, the Wavelet transform provides a time-frequency (or space-frequency) representation, allowing for localized evaluation in both space and frequency domains. 

Wavelet transform was applied to all datasets. As was done with the Fourier transform, the experiments included the classification accuracy using the full images, as well as experiments with the 20$\times$20 cropped sub-images. The Wavelets transformation were generated by using the {\it PyWavelets} Python package \citep{pywt:2024}. Two commonly used wavelets were used in this study: Haar and Daubechies. Haar wavelets are discontinuous and seem like a step function. It is a simple form of wavelet that can be used to test the impact of wavelets in the broad sense. Daubechies wavelets best represent polynomial trends, and are used often in situations of signal discontinuity, which is a situation that can be used to separate between foreground signal and detectable background ``noise''. 

\begin{figure}[!ht]
    \centering
    \includegraphics[scale=0.4]{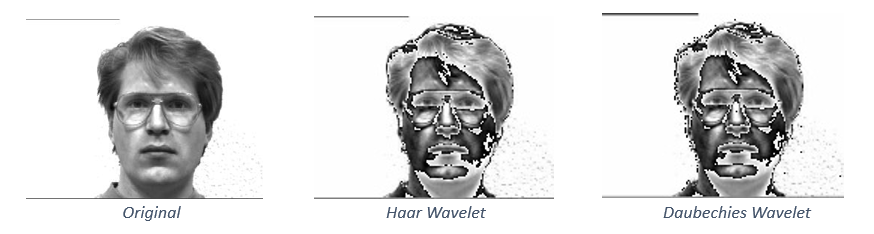}
    \caption{Example of an original image and the Haar and Daubechies Discrete Wavelet transformed images.}
    \label{fig_wavelet}
\end{figure}

\begin{figure}[!ht]
    \centering
    \includegraphics[scale=0.4]{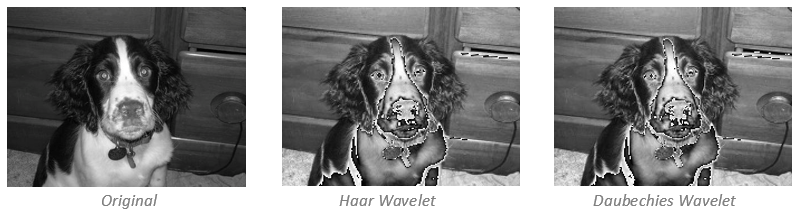}
    \caption{Example of an original image taken from ImageNette, and the Haar and Daubechies Discrete Wavelet transformed images.}
    \label{Imagenette_Wavelet}
\end{figure}

In this study, the Wavelets are implemented through the two dimensional Wavelte transform function (dwt2) of the {\it Wavelet Transforms in Python} library (PyWavelets) with the ``haar'' parameter and the ``daubechies'' parameter. The code used for the transform is available at \url{https://github.com/SaiTeja-Erukude/identifying-bias-in-dnn-classification}. Figures~\ref{fig_wavelet} and~\ref{Imagenette_Wavelet} show the images produced by applying these two wavelets. The transforms appear comparable to each other using the unaided human eye.

\begin{figure}[!ht]
    \centering
    \includegraphics[scale=0.6]{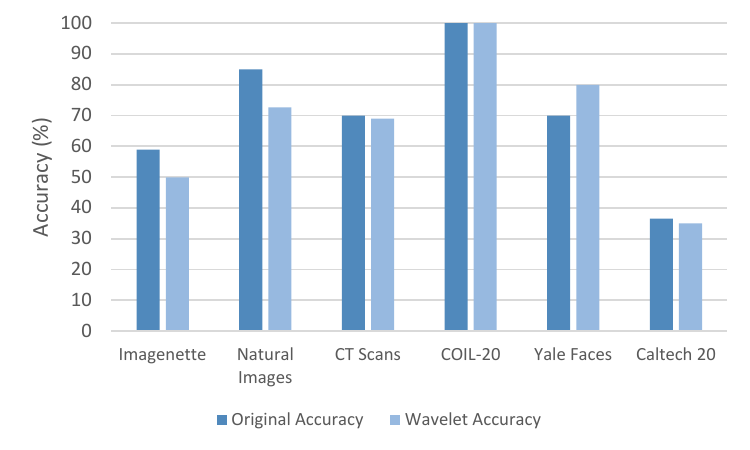}
    \caption{Classification accuracy when using images after applying the Wavelet transform.}
    \label{fig_wt_full_graph}
\end{figure}

The Wavelet findings indicate a notable decline in classification accuracy for natural datasets, while non-natural datasets have seen either consistent or improved accuracy rates. This trend is observed in both full images and cropped images. For instance, as shown in Figure~\ref{fig_wt_full_graph}, the prediction accuracy for the Imagenette dataset has dropped from $\sim$59\% to around $\sim$50\% for full images, and from $\sim$17\% to roughly $\sim$13\% (nearly random levels) for cropped images. In contrast, the COIL-20 dataset, which is synthetic, accuracy maintains a constant level of 100\%, and its cropped image accuracy has increased from $\sim$27.8\% to about $\sim$31.4\%. 

\begin{figure}[!ht]
    \centering
    \includegraphics[scale=0.6]{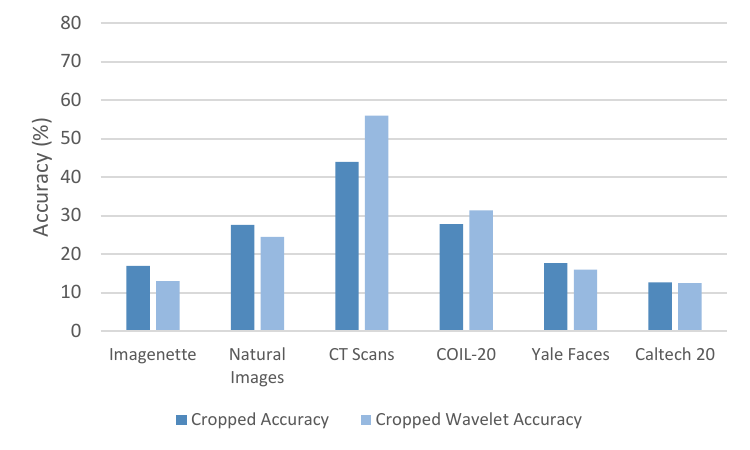}
    \caption{Classification accuracy when using the Wavelet transformed cropped images}
    \label{fig_wt_cropped_graph}
\end{figure}

The 20$\times$20 cropped images are seemingly blank and do not contain any visual information that can be sensed by the human eye. However, they still contain hidden information that can influence a CNN model's predictions. As shown in Figure~\ref{fig_wt_cropped_graph}, applying a wavelet transform to these cropped images has increased their accuracy compared to the accuracy using the original images. This indicates that the Wavelet transform revealed and enhanced the hidden signal in the background. That means that CNN has learned to use the background information, leading to more biased and therefore less reliable results.

In contrast, natural datasets made of images collected from various sources typically do not contain bias in the imaging process, as images of the same class are rarely taken during the same session or provided by the same single source. Therefore, these datasets are affected differently by the wavelet transform. As a result, the classification accuracy for Imagenette and Natural image datasets do not improve when the wavelet transform is applied. Therefore, the wavelet transform effectively differentiates between contextual information and the presence of background information that leads to bias. It can therefore be used to test for the possible presence of background information that can lead to bias, yet without confusing it with contextual information.

\subsection{Median filter}
\label{median_filter}

The median filter is a very common non-linear simple image processing method that evaluates the image pixel-by-pixel and replaces each pixel with the median of its adjacent entries. The pattern of nearby pixels depicts a window that slides across the entire image, one entry at a time \citep{DHANANI201319}. It is a smoothing technique that effectively removes disturbances from a noisy image or a signal. Unlike low-pass finite impulse response (FIR) filters, median filters are notable for retaining image edges, making them commonly used in image processing. Median filtering was performed on all datasets described in Section~\ref{data} using OpenCV's `medianBlur()` \citep{opencv_median:2024} method with windows size of 5$\times$5 to smooth the images and reduce noise. 



For full images, the classification accuracy decreased when using natural datasets with the median filter. For example, Imagenette dropped from $\sim$59\% to about $\sim$55\%, and Natural images fell from $\sim$85\% to about $\sim$81.6\%. In contrast, the accuracy for synthetic datasets has remained stable, while hybrid datasets have improved. For instance, the accuracy for Caltech 20 increased from around  $\sim$36.6\% to around $\sim$39\%. 

\begin{figure}[!ht]
    \centering
    \includegraphics[scale=0.6]{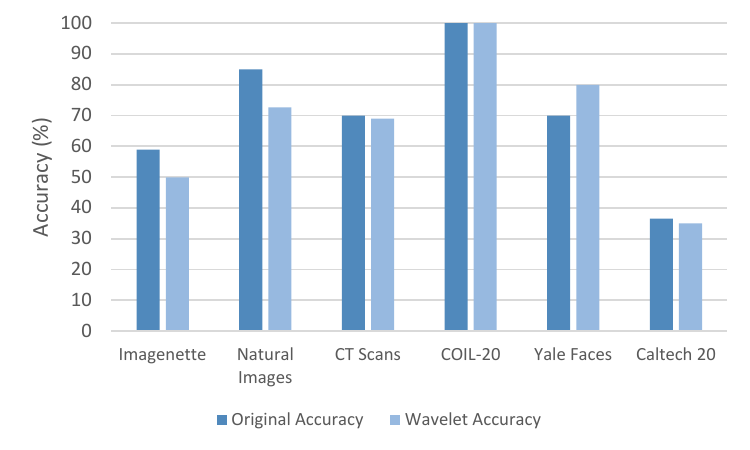}
    \caption{Classification accuracy when using the full images after applying median filtering.}
    \label{fig_median_full_graph}
\end{figure}

\begin{figure}[!ht]
    \centering
    \includegraphics[scale=0.6]{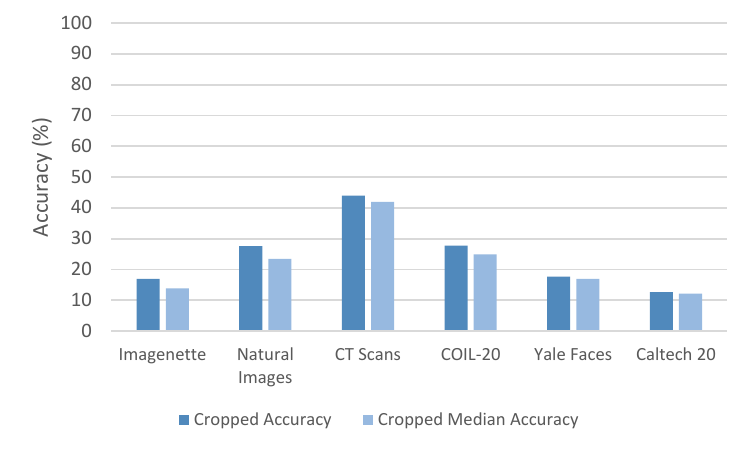}
    \caption{Classification accuracy when using the sub-images of the blank background after applying median filter.}
    \label{fig_median_cropped_graph}
\end{figure}

The median filter had a relatively small window of 5$\times$5. The size of the window does not significantly affect the classification accuracy. Only when the window is larger than 25$\times$25 the classification accuracy starts to decrease, but for smaller window sizes the results are nearly identical. For instance, window size of 7$\times$7 or 9$\times$9 can be used with no substantial impact on the results.  






The median filters were also tested in combination with the wavelet transform, such that the wavelet transform was applied after applying the median filter. As before, the analysis was applied to the full images, as well as the cropped images. The accuracy for natural datasets, such as Imagenette and Natural images, decreased by about $\sim$9.5\% and $\sim$14.8\%, respectively. The classification accuracy for synthetic and hybrid datasets either remained stable or increased. That indicates that when applying the median filter followed by a wavelet transform, the classification accuracy when using the datasets that are known to contain bias increases or remains the same, while the accuracy of datasets collected from various sources drops. 


\begin{figure}[!ht]
    \centering
    \includegraphics[scale=0.6]{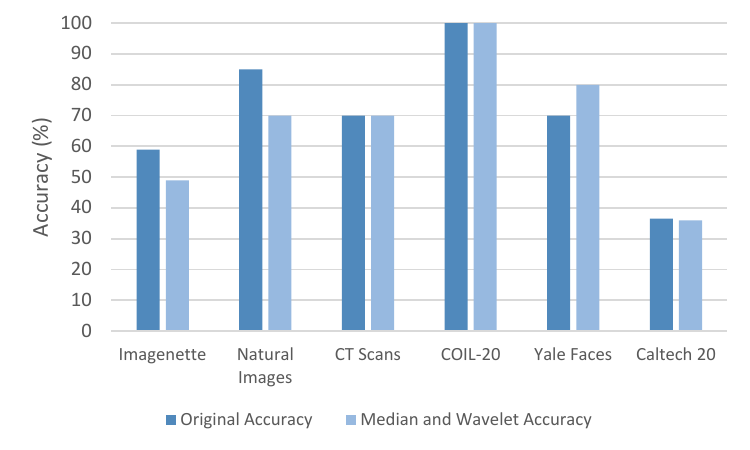}
    \caption{Classification accuracy of full images after applying both the median and wavelet transforms.}
    \label{fig_mwt_full_graph}
\end{figure}

\begin{figure}[!ht]
    \centering
    \includegraphics[scale=0.6]{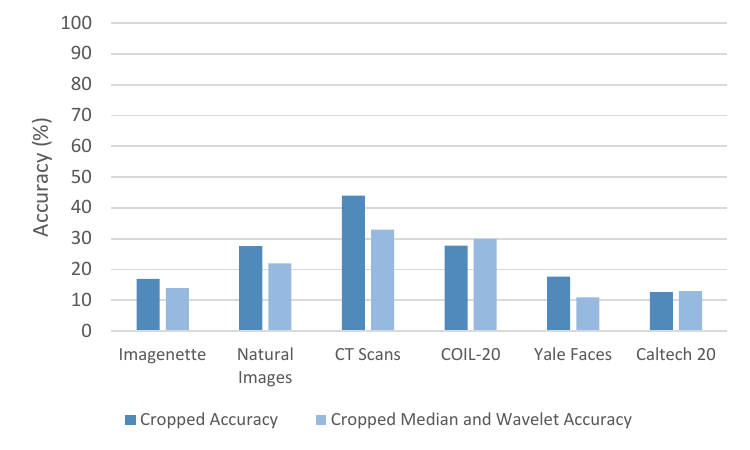}
    \caption{Classification accuracy of the blank background sub-images after applying both the median and wavelet transforms.}
    \label{fig_mwt_cropped_graph}
\end{figure}

\subsection{Summary of the results}

Table~\ref{f1} summarizes the results discussed in Section~\ref{methods_and_results}. The table shows the classification accuracy, but also the specificity, sensitivity, and F1 score. The table provides a higher resolution view of the results. The table provides a higher resolution view of the results.

\begin{table*}[]
\scriptsize
\begin{tabular}{llllll}
Dataset        & Original Images   & Fourier Transformed     & Wavelet Transformed     & Median Filtered   & Median + Wavelet                                                                                                                   \\
\hline
Imagenette     & \begin{tabular}[c]{@{}l@{}}Accuracy: 59\%\\    \\ Precision: 57.5\%\\    \\ Recall: 59.6\%\\    \\ F1: 0.5853\end{tabular}   & \begin{tabular}[c]{@{}l@{}}Accuracy: 38\%\\    \\ Precision: 37.1\%\\    \\ Recall: 38.2\%\\    \\ F1 Score: 0.3764\end{tabular}   & \begin{tabular}[c]{@{}l@{}}Accuracy: 50\%\\    \\ Precision: 48.3\%\\    \\ Recall: 49.7\%\\    \\ F1: 0.4899\end{tabular}         & \begin{tabular}[c]{@{}l@{}}Accuracy: 55\%\\    \\ Precision: 52.3\%\\    \\ Recall: 53.4\%\\    \\ F1: 0.5284\end{tabular}       & \begin{tabular}[c]{@{}l@{}}Accuracy: 49.5\%\\    \\ Precision: 47\%\\    \\ Recall: 50\%\\    \\ F1: 0.4845\end{tabular}           \\
\hline
Natural Images & \begin{tabular}[c]{@{}l@{}}Accuracy: 85\%\\    \\ Precision: 86\%\\    \\ Recall: 85.33\%\\    \\ F1: 0.8552\end{tabular}    & \begin{tabular}[c]{@{}l@{}}Accuracy: 76.8\%\\    \\ Precision: 77.3\%\\    \\ Recall: 76.6\%\\    \\ F1 Score: 0.7660\end{tabular} & \begin{tabular}[c]{@{}l@{}}Accuracy: 72.8\%\\    \\ Precision: 73.7\%\\    \\ Recall: 73.5\%\\    \\ F1 Score: 0.7261\end{tabular} & \begin{tabular}[c]{@{}l@{}}Accuracy: 81.6\%\\    \\ Precision: 82\%\\    \\ Recall: 81.3\%\\    \\ F1 Score: 0.8157\end{tabular} & \begin{tabular}[c]{@{}l@{}}Accuracy: 70.6\%\\    \\ Precision: 72.7\%\\    \\ Recall: 70.7\%\\    \\ F1 Score: 0.7134\end{tabular} \\
\hline
CT Scans       & \begin{tabular}[c]{@{}l@{}}Accuracy: 70\%\\    \\ Precision: 66.7\%\\    \\ Recall: 70.6\%\\    \\ F1: 0.6859\end{tabular}   & \begin{tabular}[c]{@{}l@{}}Accuracy: 54\%\\    \\ Precision: 52.1\%\\    \\ Recall: 54.4\%\\    \\ F1 Score: 0.5323\end{tabular}   & \begin{tabular}[c]{@{}l@{}}Accuracy: 69\%\\    \\ Precision: 67.2\%\\    \\ Recall: 69.1\%\\    \\ F1: 0.6814\end{tabular}         & \begin{tabular}[c]{@{}l@{}}Accuracy: 69\%\\    \\ Precision: 67\%\\    \\ Recall: 68.8\%\\    \\ F1: 0.6789\end{tabular}         & \begin{tabular}[c]{@{}l@{}}Accuracy: 70.5\%\\    \\ Precision: 67.3\%\\    \\ Recall: 70.9\%\\    \\ F1: 0.6905\end{tabular}       \\
\hline
COIL-20        & \begin{tabular}[c]{@{}l@{}}Accuracy: 100\%\\    \\ Precision: 100\%\\    \\ Recall: 100\%\\    \\ F1: 1\end{tabular}         & \begin{tabular}[c]{@{}l@{}}Accuracy: 80\%\\    \\ Precision: 81.2\%\\    \\ Recall: 80\%\\    \\ F1 Score: 0.8060\end{tabular}     & \begin{tabular}[c]{@{}l@{}}Accuracy: 100\%\\    \\ Precision: 100\%\\    \\ Recall: 100\%\\    \\ F1: 1\end{tabular}               & \begin{tabular}[c]{@{}l@{}}Accuracy: 100\%\\    \\ Precision: 100\%\\    \\ Recall: 100\%\\    \\ F1: 1\end{tabular}             & \begin{tabular}[c]{@{}l@{}}Accuracy: 100\%\\    \\ Precision: 100\%\\    \\ Recall: 100\%\\    \\ F1: 1\end{tabular}               \\
\hline
Yale Faces     & \begin{tabular}[c]{@{}l@{}}Accuracy: 70\%\\    \\ Precision: 75.4\%\\    \\ Recall: 70\%\\    \\ F1: 0.7260\end{tabular}     & \begin{tabular}[c]{@{}l@{}}Accuracy: 60\%\\    \\ Precision: 63.8\%\\    \\ Recall: 60.4\%\\    \\ F1 Score: 0.6205\end{tabular}   & \begin{tabular}[c]{@{}l@{}}Accuracy: 80\%\\    \\ Precision: 82.2\%\\    \\ Recall: 80.1\%\\    \\ F1: 0.8114\end{tabular}         & \begin{tabular}[c]{@{}l@{}}Accuracy: 70\%\\    \\ Precision: 74.1\%\\    \\ Recall: 71.3\%\\    \\ F1: 0.7267\end{tabular}       & \begin{tabular}[c]{@{}l@{}}Accuracy: 80\%\\    \\ Precision: 83\%\\    \\ Recall: 80.7\%\\    \\ F1: 0.8183\end{tabular}           \\
\hline
Caltech 20     & \begin{tabular}[c]{@{}l@{}}Accuracy: 36.6\%\\    \\ Precision: 30.4\%\\    \\ Recall: 35.2\%\\    \\ F1: 0.3262\end{tabular} & \begin{tabular}[c]{@{}l@{}}Accuracy: 30\%\\    \\ Precision: 25.3\%\\    \\ Recall: 29.1\%\\    \\ F1: 0.2707\end{tabular}         & \begin{tabular}[c]{@{}l@{}}Accuracy: 35\%\\    \\ Precision: 33.3\%\\    \\ Recall: 34.8\%\\    \\ F1: 0.3403\end{tabular}         & \begin{tabular}[c]{@{}l@{}}Accuracy: 39\%\\    \\ Precision: 37.1\%\\    \\ Recall: 38.6\%\\    \\ F1: 0.3784\end{tabular}       & \begin{tabular}[c]{@{}l@{}}Accuracy: 36\%\\    \\ Precision: 32\%\\    \\ Recall: 35\%\\    \\ F1: 0.3343
\end{tabular}    
\end{tabular}
\caption{The classification accuracy, F1 scores, precision and recall of the experiments.}
\label{f1}
\end{table*}

\section{Conclusion}
\label{conclusion}

CNNs have become the most prevalent computational tool for image classification. 
Due to its black-box nature, reliable performance evaluation can become challenging, and biases or unreliable performance evaluations can go unnoticed. 
In this study, several techniques including Fourier transforms, Wavelet transforms, simple median filtering, and their combinations were tested. The Wavelet transform, simple median filtering transform, and their combination have shown to be effective in distinguishing between biases in natural datasets that is driven by contextual information, and bias in non-natural datasets that have been shown in previous studies to be driven by artifacts acquired through the imaging process. Therefore, the results show that the method can distinguish between contextual information and imaging process bias, and alert on the presence of such bias even without the need to separate background information from the original images.

Applying a median filter smooths the image and eliminates noise, and subsequently applying wavelet transform to this smoothed dataset further enhances the signal in the image. That process affects the contextual visual information in a different manner than it affects the artifacts acquired through the imaging process. The wavelet transform highlights any hidden signals introduced during the image acquisition process. In summary, these methods decreased the prediction accuracy for natural datasets but improved the accuracy for non-natural datasets, effectively addressing the distinct biases present in each.

Such method should be applied to verify the soundness of CNNs when they are used to classify image datasets, and specifically datasets acquired through controlled environment such medical images. Once the neural network is trained, it is tested by classifying the images after pre-processing them as described in Section~\ref{methods_and_results}. That can provide an indication of whether bias is present. That is needed only in cases where no parts of the background that are clearly irrelevant to the image classification task are available, as the classification of such 20$\times$20 blank parts of the images can provide a simpler indication of the presence of bias.

Testing for bias using the proposed method is straightforward: First, the images are transformed using the image transformations shown in Section~\ref{methods_and_results}. The transformed images are generated from the original full size images, and not the 20$\times$20 background sub-images, as such background sub-images may not be available for every dataset. After transforming the images, the CNN is applied to train and test the classification accuracy with the transformed images. If the classification accuracy with the transformed images is similar or higher than the classification accuracy when using the original images, it is an indication that the classification might be driven by bias.

It must be reminded that the method can identify the presence of bias that is difficult to notice, but biases can have different forms and reasons. Therefore, if the proposed method does not identify the presence of bias, that does not necessarily ensure that no bias exists. Other biases can also be present, some of them might be unknown or unexpected, and the proposed method is not designed to identify all possible biases. 

The proposed method also does not correct for the bias, but merely identifies its existence. As demonstrated in Section~\ref{classification_bias}, numerous experiments were done without noticing the bias, and many commonly used benchmark datasets can be classified with high classification accuracy by just using a seemingly blank sub-image of the background. Therefore, before fully trusting the prediction accuracy of CNNs, it is recommended to apply methods such as the tests outlined in Section~\ref{methods_and_results} to identify possible biases in the datasets and it should also help you distinguish between contextual information and the presence of background bias. This evaluation helps gauge CNN's reliability and ensures its accuracy is trustworthiness.

Future work will focus on identifying the sources of the biases, and potentially altering the images to correct for biases. Correction of the biases can make use of GANs or adversarial neural networks creating images that visually seem identical to the original images, but are handled differently by deep neural networks. Such pre-processing can not merely identify the bias, but also help to avoiding it by removing information that is irrelevant to the problem at hand but can still be used by classifiers to classify the images.   

\section*{Code availability}

The complete Python code utilized in this research study is available in a GitHub repository:
\url{https://github.com/SaiTeja-Erukude/identifying-bias-in-dnn-classification}. Using the code will require to change the input and output directories. Required libraries are pywt, cv2, and keras.


\section*{Acknowledgments}

This study was supported in part by NSF grant 2148878.

\bibliographystyle{apalike}
\bibliography{main_arxiv}

\end{document}